\documentclass{article}

\usepackage[final]{neurips_bdl2019}
\usepackage{fontawesome}

\newcommand{\EO}{\mathrm{EL}_2\mathrm{O}}
\def\bi#1{\hbox{\boldmath{$#1$}}}

\usepackage[utf8]{inputenc} % allow utf-8 input
\usepackage[T1]{fontenc}    % use 8-bit T1 fonts
\usepackage{hyperref}       % hyperlinks
\usepackage{url}            % simple URL typesetting
\usepackage{booktabs}       % professional-quality tables
\usepackage{amsfonts}       % blackboard math symbols
\usepackage{nicefrac}       % compact symbols for 1/2, etc.
\usepackage{microtype}      % microtypography
\usepackage{aas_macros}
\usepackage{amsmath}
\usepackage{graphicx}
\usepackage{subcaption}
\usepackage{wrapfig}
\usepackage{floatrow}

\title{Uncertainty Quantification with Generative Models}

\author{Vanessa B\"ohm, Fran\c{c}ois Lanusse, Uro\v s Seljak\\
Berkeley Center for Cosmological Physics, Department of Physics\\
University of California, Berkeley, California, USA\\ 
Lawrence Berkeley National Laboratory, Cyclotron Rd, Berkeley, California, USA\\
}

\begin{document}

\maketitle

\begin{abstract}
%   The abstract paragraph should be indented \nicefrac{1}{2}~inch (3~picas) on
%   both the left- and right-hand margins. Use 10~point type, with a vertical
%   spacing (leading) of 11~points.  The word \textbf{Abstract} must be centered,
%   bold, and in point size 12. Two line spaces precede the abstract. The abstract
%   must be limited to one paragraph.
We develop a generative model-based approach to Bayesian inverse problems, such as image reconstruction from noisy and incomplete images. Our framework addresses two common challenges of Bayesian reconstructions: 
1) It makes use of complex, data-driven priors that comprise all available information about the uncorrupted data distribution.
2) It enables computationally tractable uncertainty quantification in the form of posterior analysis in latent and data space. The method is very efficient in that the generative model only has to be trained once on an uncorrupted data set, after that, the procedure can be used for arbitrary corruption types.
\end{abstract}

\section{Introduction}
Data reconstruction from corrupted data lies at the heart of Bayesian inverse problems. There are many ways in which data can be corrupted, ranging from missing data (or masking) to noise and blurring. In many cases the reconstructed data, e.g. an image, is high dimensional. Data reconstruction commonly faces two challenges: 1) For most data it is difficult to find a prior that optimally represents our knowledge of the uncorrupted data. Instead, priors are often chosen to impose certain regularity conditions (e.g. maximum entropy, smoothness). 2) While it is usually tractable to find the maximum of the posterior, i.e. the most probable underlying realization, a full uncertainty quantification of the reconstruction is usually prohibitively expensive.

We propose to address these challenges with generative models, which provide a mapping from points in a typically lower dimensional latent space to points in the high dimensional data space. As an example of such a model we will be using a Variational AutoEncoder (VAE) \citep{KingmaWelling13,RezendeMW14}. However, the use of a VAE is not a limitation of this method and other generative models could be used instead. VAEs are designed to model the distribution $p_{\phi}(\bi x)$ of high-dimensional input data, $\bi x$, by introducing a mapping $p_{\phi}(\bi{x} | \bi{z})$ to a lower dimensional latent representation, $\bi z$. The latent space variables are enforced to follow a given prior distribution, $p(\bi{z})$, which is typically chosen to be a standard normal distribution.\footnote{Other generative models might not fulfill this condition of a Gaussian latent space distribution. For our method to work we only require the distribution of latent space variables to be fairly well behaved such that it can be mapped to a Gaussian, e.g. with a bijective normalizing flow, as we explain later.}

Given a generative model, the posterior of the latent variables for a given data realization can be modeled with Bayes rule
\begin{equation}
\label{eq:1}
       p_{\phi}(\bi{z} | \bi{x}) \propto p_{\phi}(\bi{x} | \bi{z}) p(\bi{z}).
\end{equation}
This formulation allows to address both aforementioned problems: 1) The prior distribution, $p(\bi{z})$, reflects the distribution of the training data. 2) The representation of the posterior in the lower dimensional latent space enables tractable posterior analysis. In particular it allows to examine and fit the posterior distribution and draw samples from it. The samples can then be visualized in data space by forward modeling with the generative model. This approach to uncertainty quantification is in the spirit of commonly applied sampling approaches like Hamiltonian Monte Carlo (HMC) sampling. In fact, it is the only way to reliably quantify uncertainty in high dimensional spaces, where often already the covariance is too big to fit into memory. Quantifying uncertainty by means of the variance instead of the covariance can be misleading in the presence of strong correlations between pixels, which are typically expected for image data. In addition, neither variance nor covariance can be used to fully characterize non-Gaussian distributions such as the multimodal distributions, which are common for corrupted data. All of these properties (complex correlations, multimodality) are captured in samples drawn from a sufficiently accurate fit to the posterior distribution in latent space. 

In addition to addressing the problems of accurate priors and tractable posterior analysis, the suggested approach is very efficient, because the generative model only has to be trained once on uncorrupted data and can then be used for arbitrary types of data corruption.
\section{Reconstruction and Uncertainty Quantification for Linear Inverse Problems}
In this work we consider linear inverse problems of the form
%\begin{equation}
$\bi{y} = \mathbf{A} \bi{x} + \bi{n}$,
%    \label{eq:problem_description}
%\end{equation}
where $\bi{x}$ is the signal to be recovered, $\bi{y}$ are the observations,  $\bi{n} \sim \mathcal{N}(0, \bi{\sigma}^2_{\textrm{data}})$ represents Gaussian noise, and $\mathbf{A}$ is a corruption operator (i.e. a masking operator for an inpainting problem, or a convolution by a blurring kernel for a deconvolution problem). 

In our Bayesian approach, we assume that the uncorrupted signal, $\bi{x}$, is drawn from a latent variable model, $p_\phi( \bi{x} | \bi{z}) p(\bi{z})$, with normal Gaussian prior $p(\bi{z}) = \mathcal{N}(0, I)$. While different likelihood models could be used to describe $p_\phi(\bi{x} |\bi{z})$, we adopt a Gaussian model $p_\phi(\bi{x} | \bi{z}) = \mathcal{N}(g_\phi(\bi{z}), \sigma_{\mathrm{model}}^2)$, where $g_\phi(\bi{z})$ is parameterized by a deep neural network. The variance $\sigma_{\mathrm{model}}^2$ measures the modeling error, i.e. the average discrepancy between an input image and its reconstruction after encoding and decoding with the generative model. To get an estimate of the size of this modeling error we suggest to measure its value after having trained the generative model or to leave it as a free, trainable parameter during the training. The Gaussian choice for the likelihood of the generative model allows us to easily combine it with the probability distribution of the noise in the corrupted data. We train the generative model on uncorrupted data, $\bi{x}$. Here, we use a vanilla VAE with mean-field Gaussian posterior and train it under the Evidence Lower BOund (ELBO) \citep{KingmaWelling13}.

Once the generative model is trained, we can use it to expand the inverse problem formulation $\bi{y} = \mathbf{A} \left[ g_\phi(\bi{z}) + \bi{n}_{\textrm{model}} \right] + \bi{n}$,
where we have replaced the signal, $\bi{x}$, by its generative process and added $\bi{n}_{\textrm{model}}$ which accounts for the generator's modeling error. Contrary to the previous formulation, we have now stated the problem in terms of a variable, $\bi{z}$, for which we have a simple and adequate prior (i.e. a normal Gaussian), allowing us to apply the Bayesian approach $\ln p(\bi{z}|\bi{y})=\ln p(\bi{z})+\ln p_{{\phi}}(\bi{y}|\bi{z})-\ln p(\bi{y})$, where $p(\bi{z}|\bi{y})$ is the posterior distribution of the latent variables for a given observation $\bi{y}$. Note that this formulation does not use the encoder of the generative model nor does it depend on the approximate posterior used in the VAE training.

To provide a concrete example, let us consider a problem combining missing data and Gaussian noise, for which $\textbf{A}$ is a binary masking matrix with entries equal to 1 if data is observed, 0 otherwise. The expression for the log posterior becomes,
\begin{equation}
\label{eq:2}
\ln p(\bi{z}|\bi{y})=-\sum_{i,\, {\rm unmasked}}\frac{[g_{\phi}(\bi{z})_i-y_i]^2}{2\sigma_{x,ii}^2}-\sum_j \frac{z_j^2}{2}  + \mathrm{constants},
\end{equation}
where $\sigma_{x,ii}^{-2}=\sigma^{-2}_{\mathrm{model},ii}+\sigma^{-2}_{\mathrm{data},ii}$. Note that this expression would change for different corruption types, while the generative model would remain the same, i.e. it does not need to be retrained.

Despite the dimensionality reduction, the analysis of this posterior is still challenging: 1) It is often multi-modal, because multiple solutions agree with the data. 2) Even in low dimensions, standard methods such as MCMC can be prohibitively slow and may give incorrect results in multi-modal situations \citep{WuDomkeEtAl18}. Here, we adopt the EL$_2$O method~\citep{SeljakYu19} to address these challenges. EL$_2$O replaces the true posterior, $p(\bi{z}|\bi{y})$ by a parameterized approximation, $q_\theta(\bi{z})$, and finds the optimal parameters, $\theta$, by minimizing the L$_2$ f-divergence between the two distributions. This approach needs few sampling points, does not suffer from sampling noise, and can be orders of magnitude faster than stochastic variational inference (SVI)~\citep{SeljakYu19}. 

In our examples, we use a mixture of multivariate Gaussians (GMM) as posterior approximation. Under the assumption of well separated mixture components, the EL$_2$O procedure for a GMM involves the following 3 steps: 1) finding all relevant minima in the posterior by optimization, 2) fitting Gaussians around these minima by using the Laplace approximation, 3) improving the probability distribution beyond the Gaussian approximation, if needed, 4) finding the relative weights of these mixture components. Details on the EL$_2$O procedure including equations are provided in Appendix~\ref{app:el2o}. While in principle SVI could be used instead of the EL$_2$O procedure, we find in our examples that EL$_2$O gives more accurate results and requires less tuning, while SVI is prone to numerical instabilities when fitting a full rank Gaussian (see Appendix~\ref{app:elosvi} for a direct comparison on an example). 
\section{Experiments}
\begin{figure}[h]
    \begin{subfigure}{0.3\textwidth}
    ~\vfill

    \includegraphics[width=0.23\textwidth]{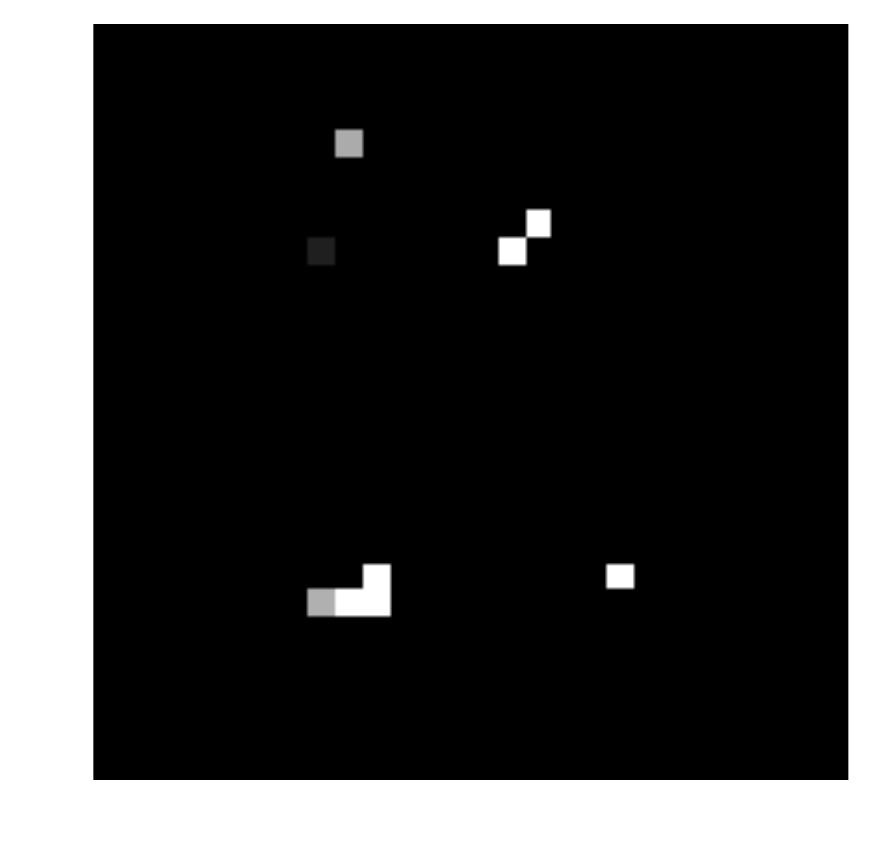}
    \includegraphics[width=0.23\textwidth]{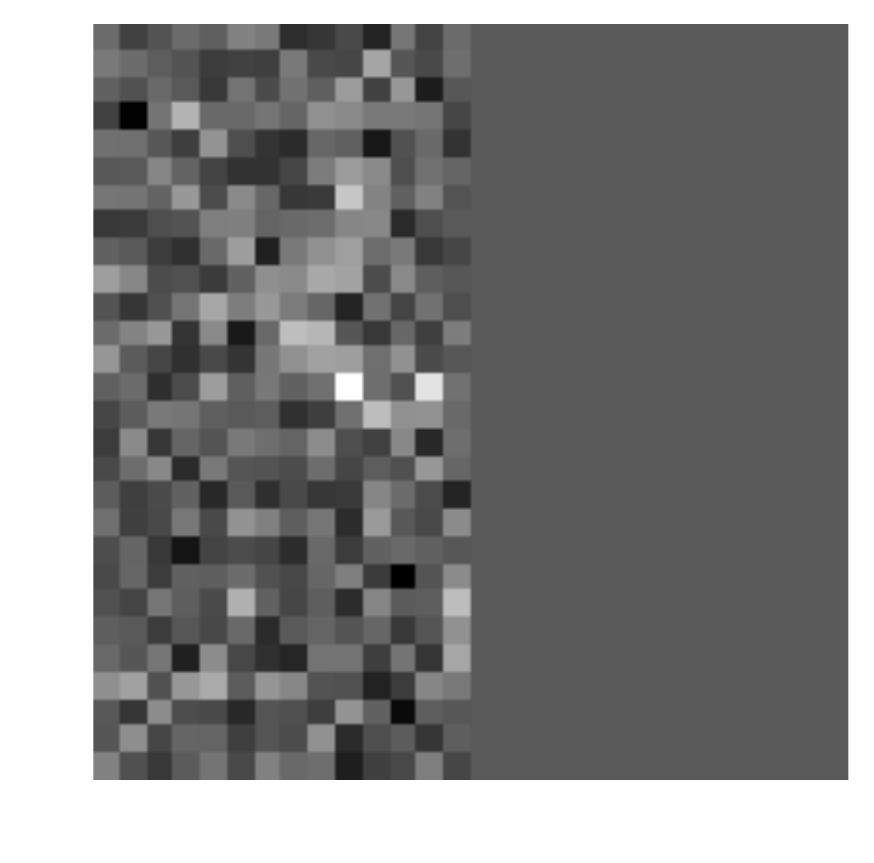}
    \includegraphics[width=0.23\textwidth]{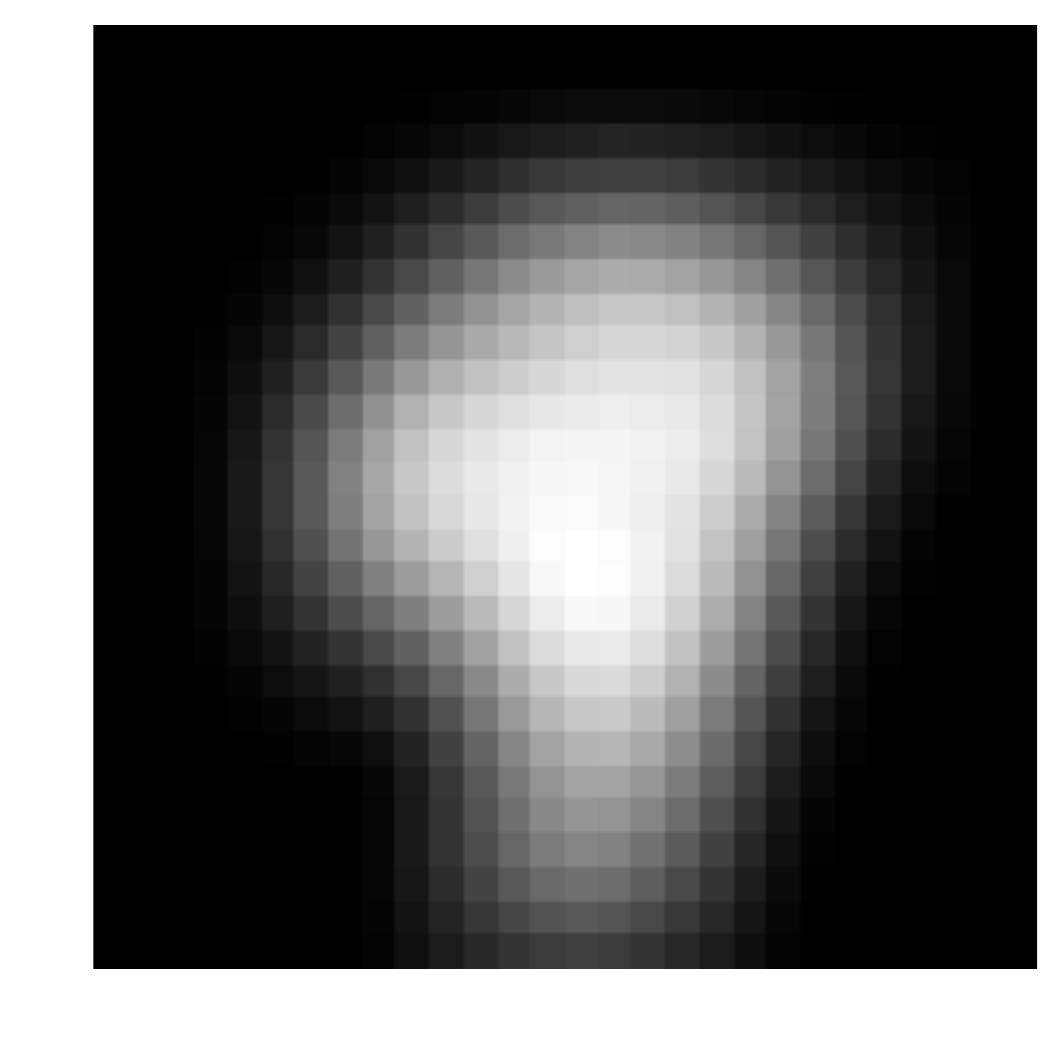}
    \includegraphics[width=0.23\textwidth]{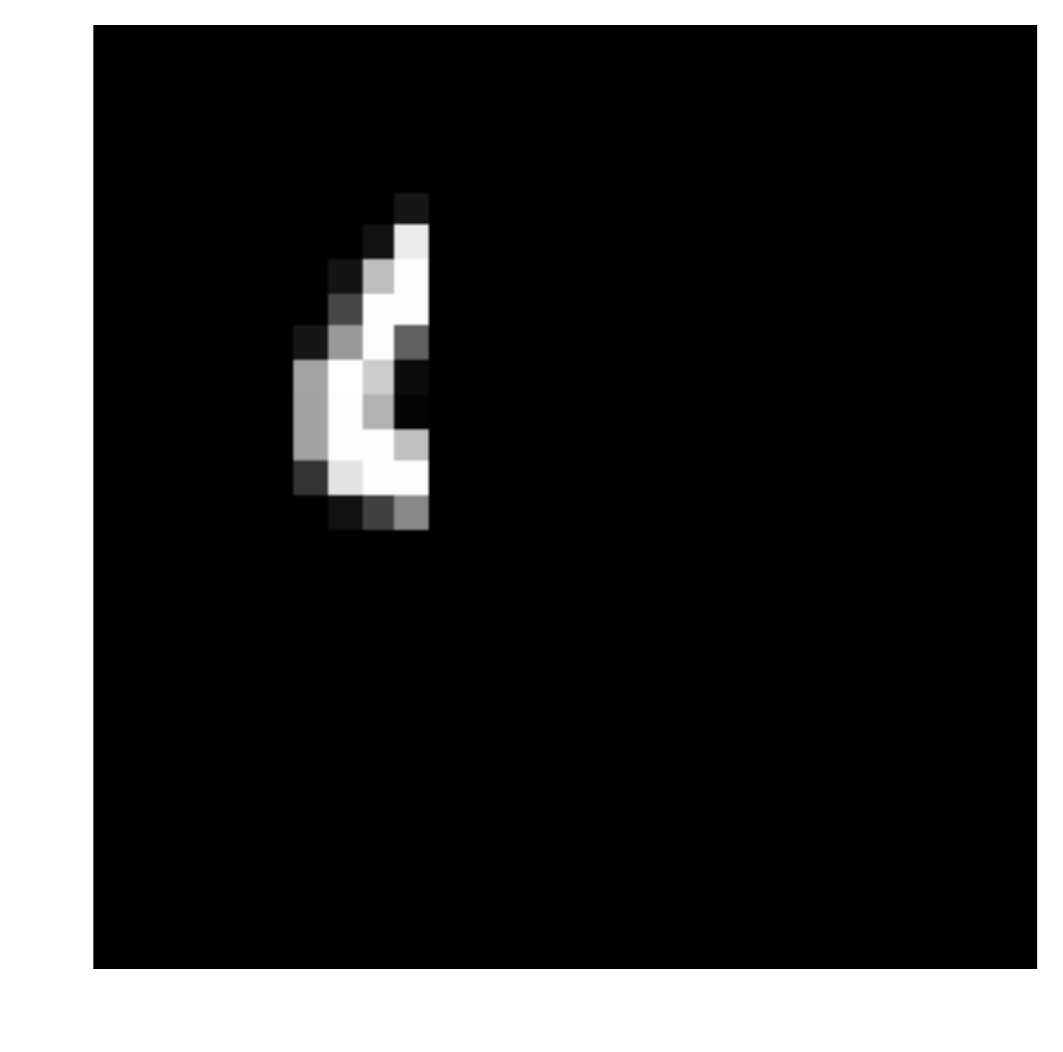}
    
    \includegraphics[width=0.23\textwidth]{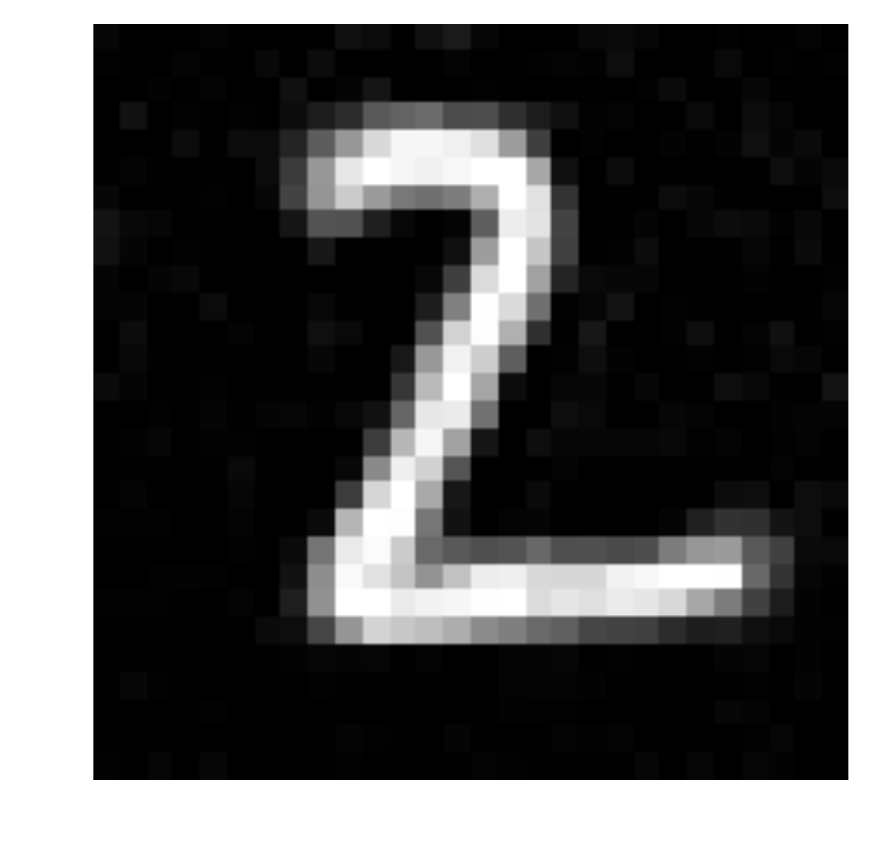}
    \includegraphics[width=0.23\textwidth]{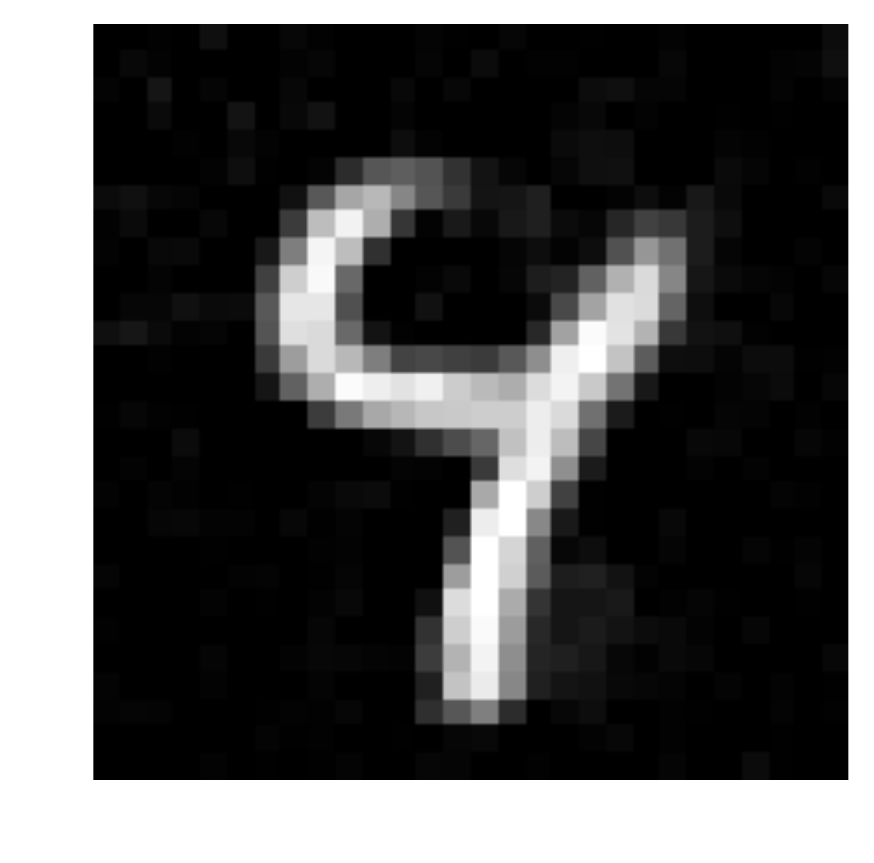}
    \includegraphics[width=0.23\textwidth]{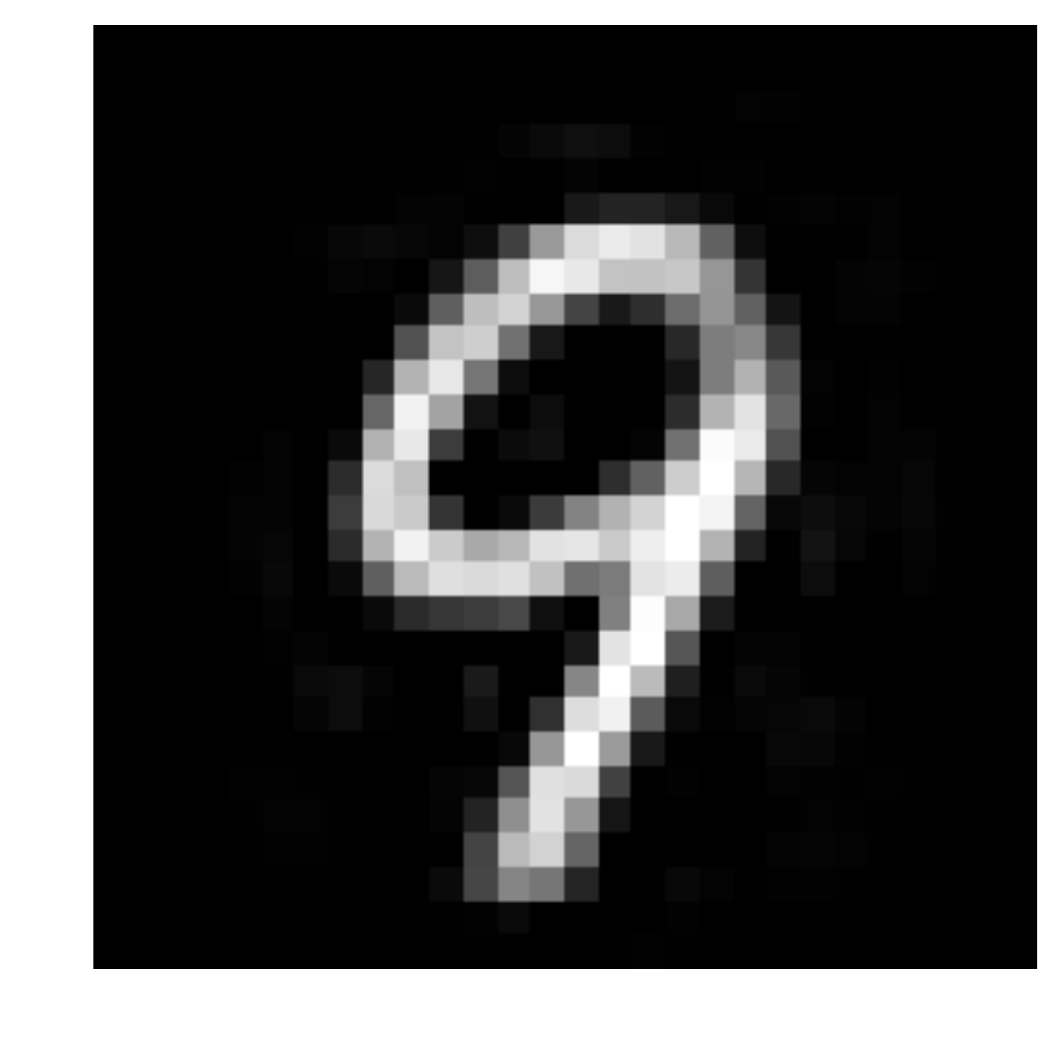}
    \includegraphics[width=0.23\textwidth]{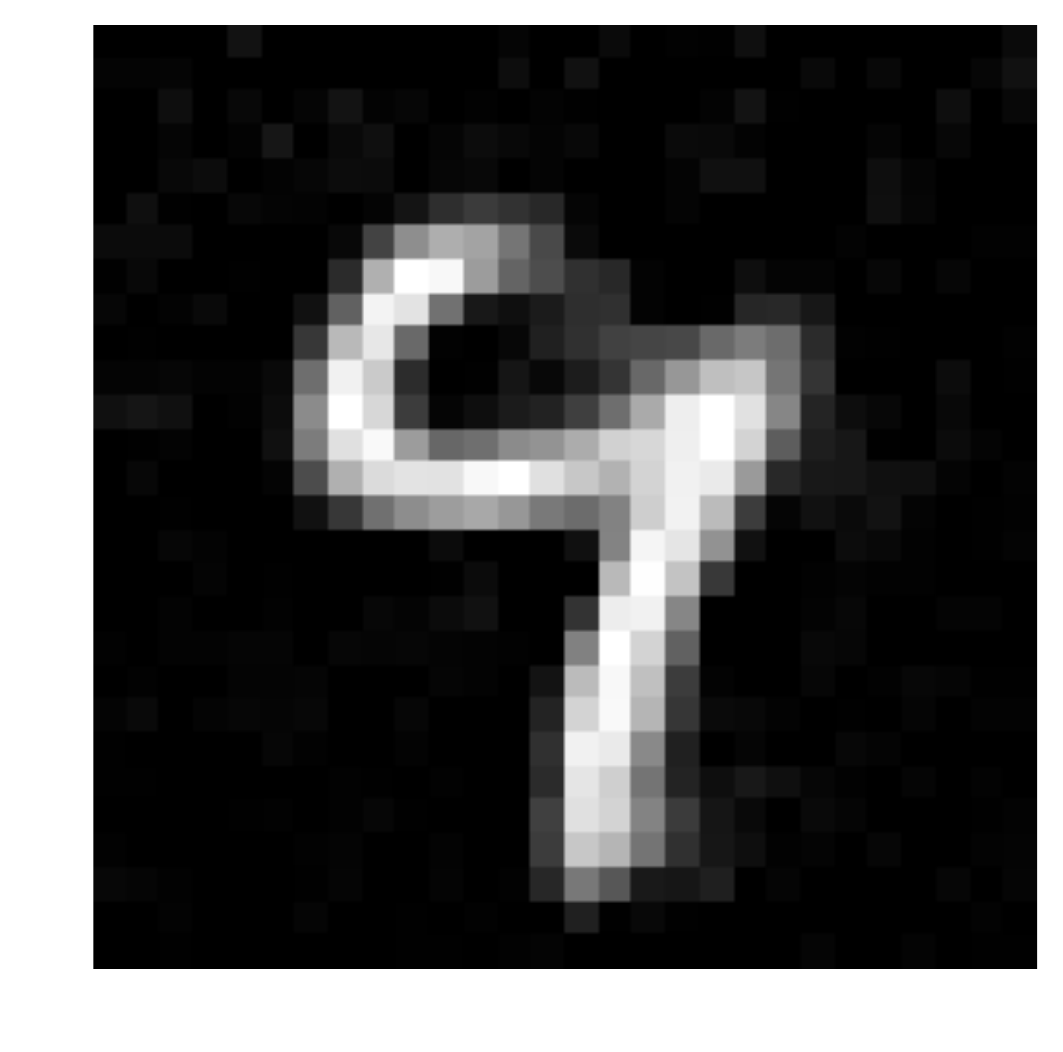}
    
    \includegraphics[width=0.23\textwidth]{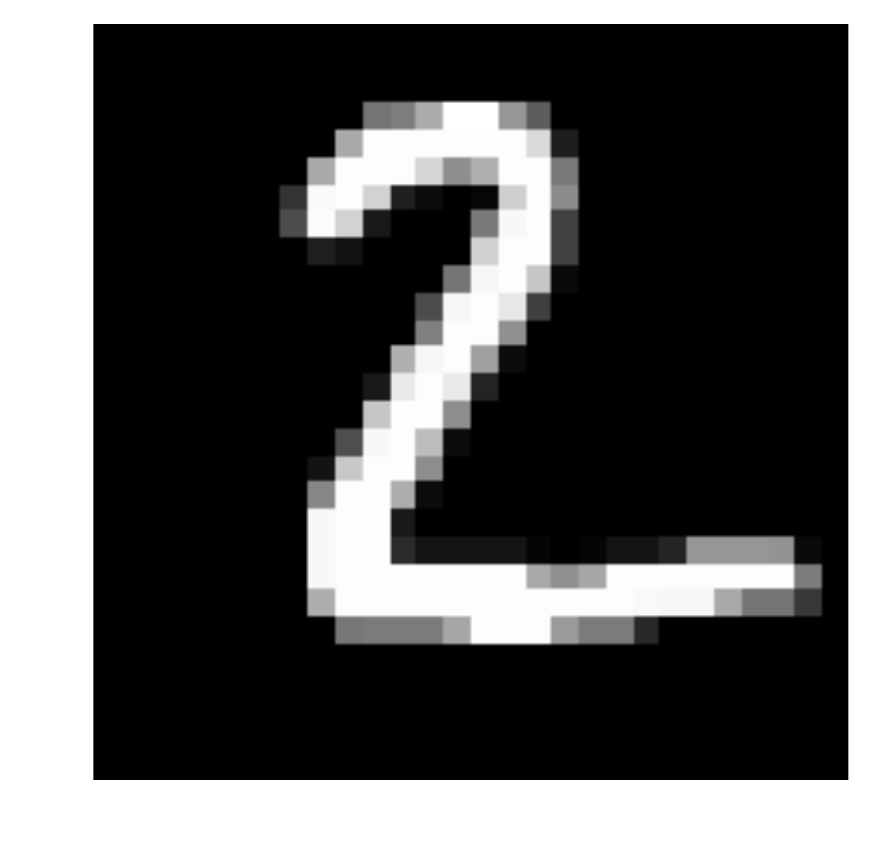}
    \includegraphics[width=0.23\textwidth]{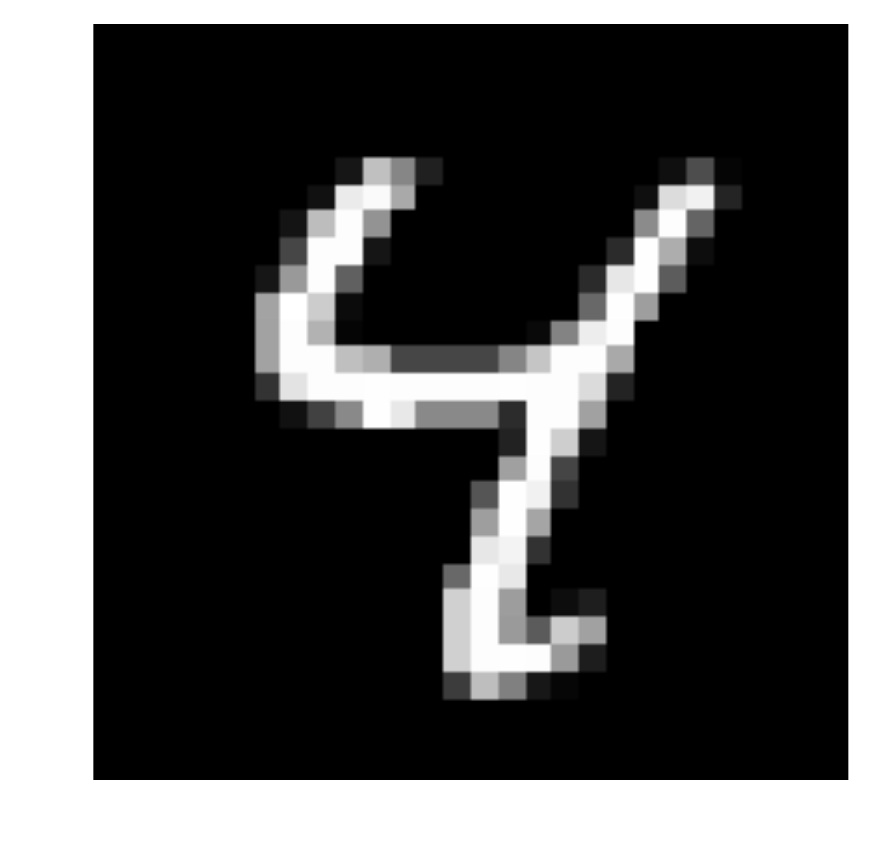}
    \includegraphics[width=0.23\textwidth]{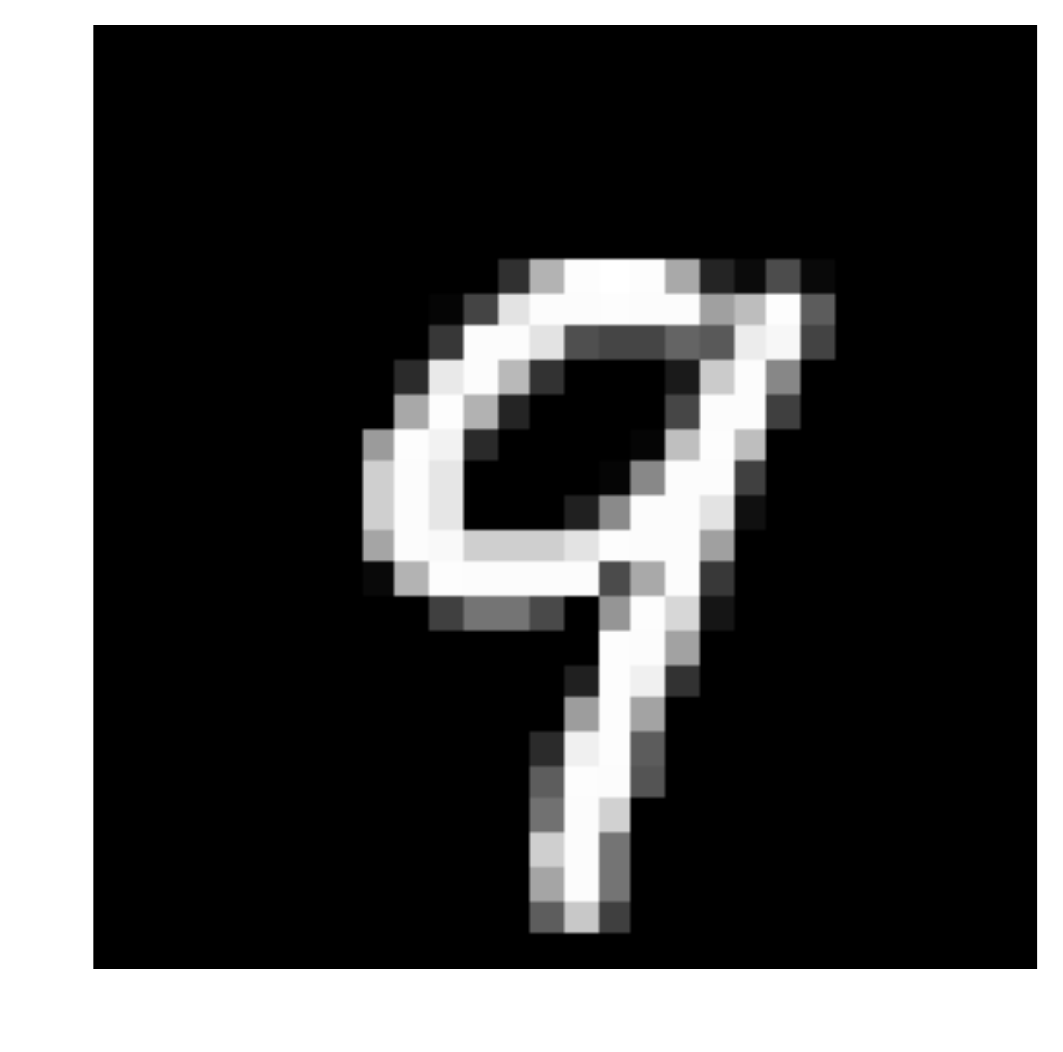}
    \includegraphics[width=0.23\textwidth]{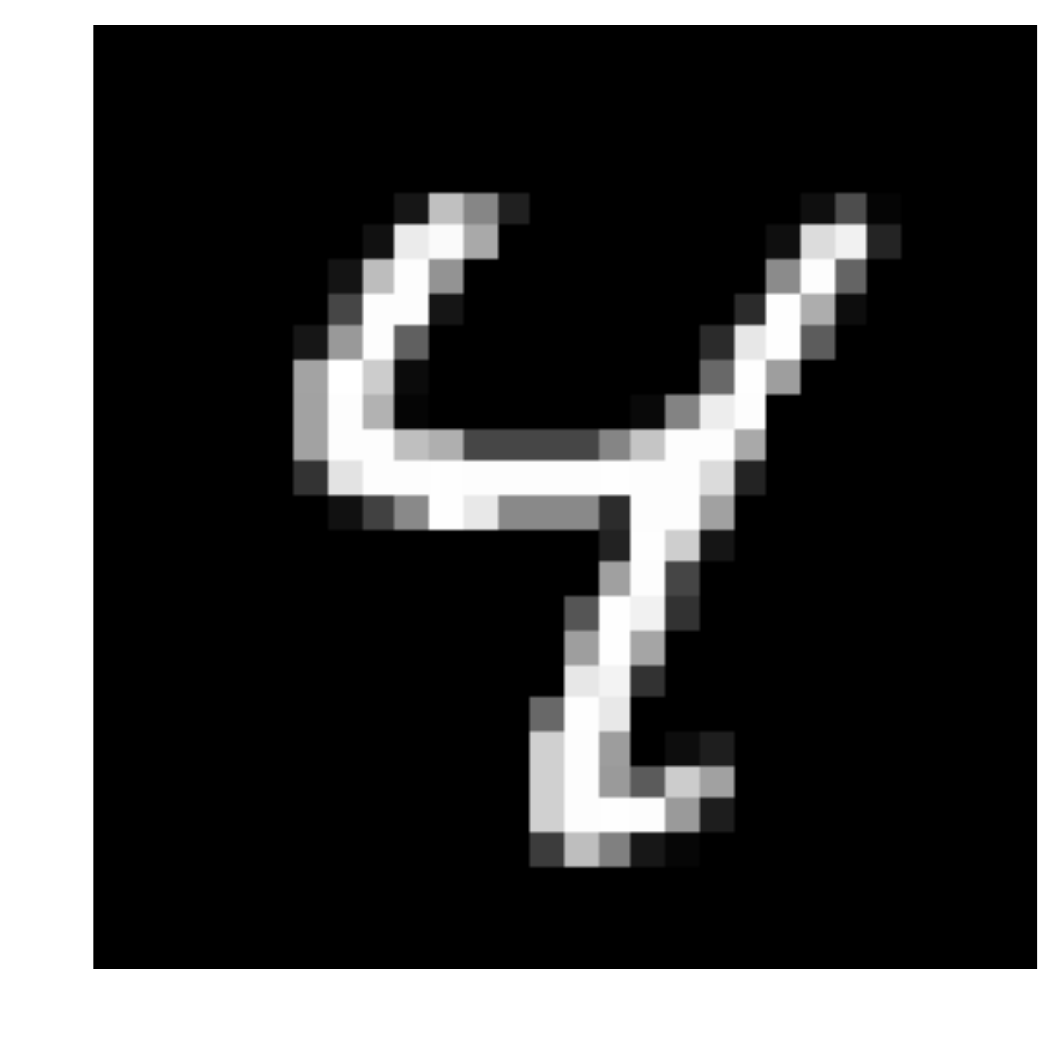}
    
    \end{subfigure}
    \begin{subfigure}{0.59\textwidth}
    \includegraphics[width=0.45\textwidth]{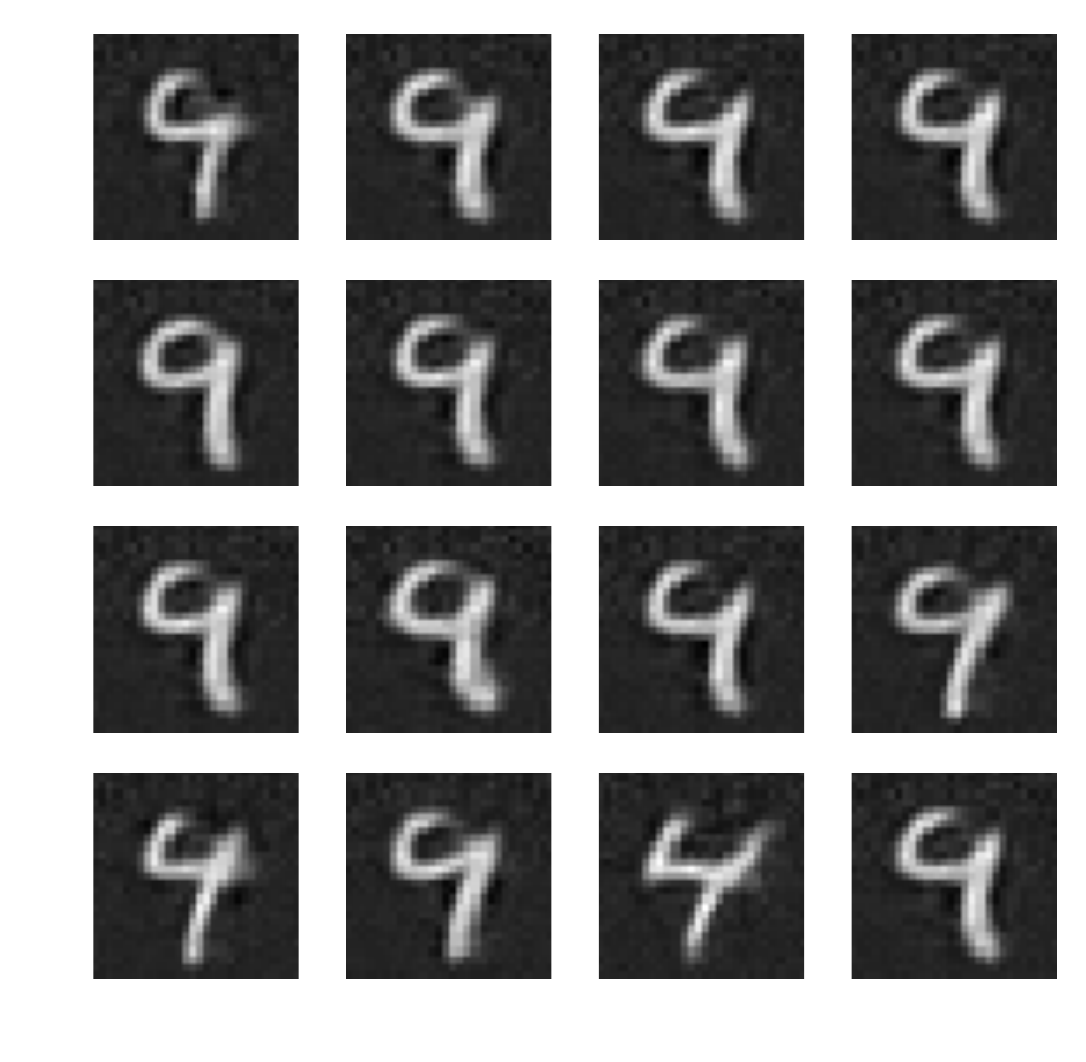}
    \includegraphics[width=0.45\textwidth]{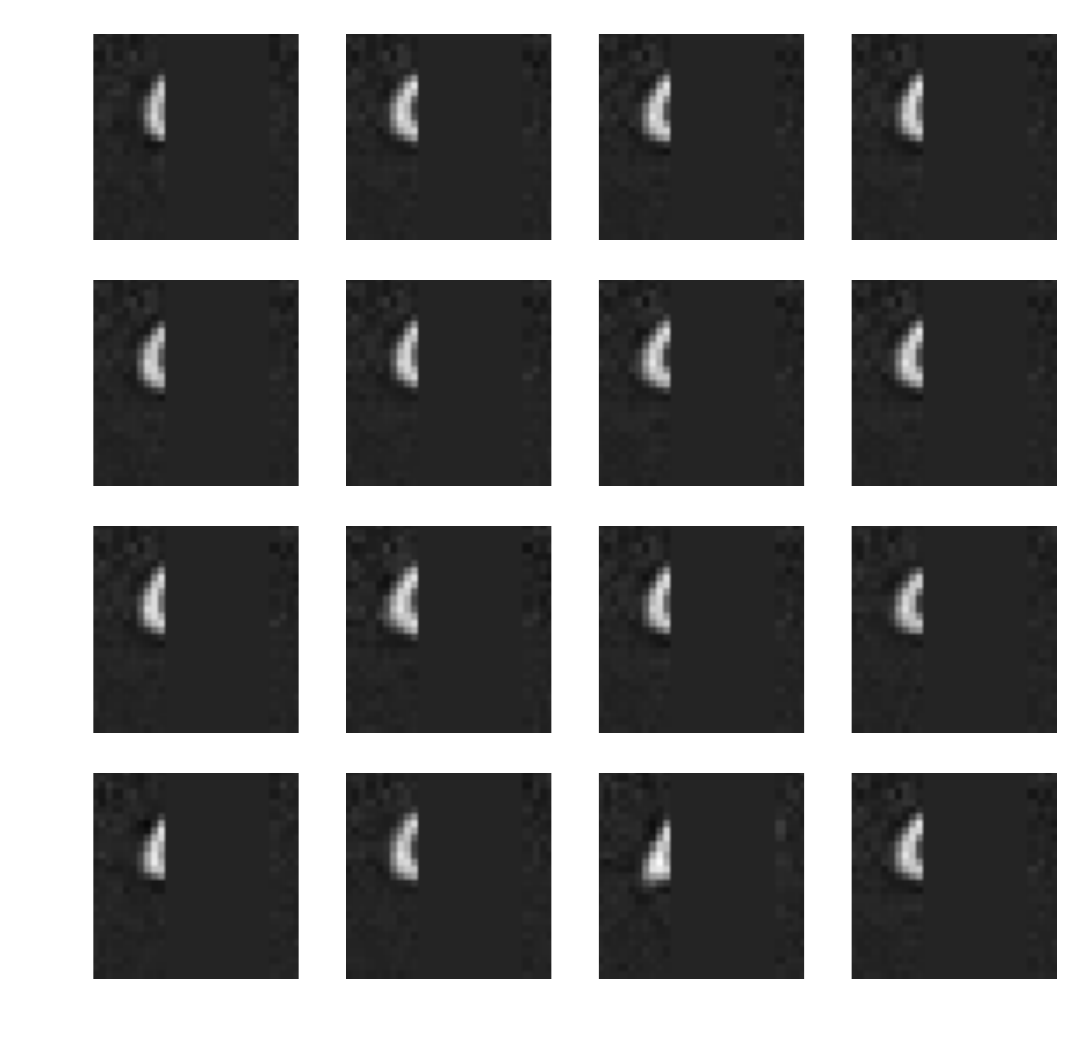}
    \end{subfigure}
    \vspace*{-4mm}
    \caption{Left panel: Data imputation, denoising and deblurring examples. From top to bottom row we show corrupted data, reconstructions obtained from forward modeling the latent space variables at the deepest minimum of the negative log posterior (the MAP solutions) and the underlying true images. Middle panel: Forward modeled samples of the example in the last column of the left panel (drawn from the two component Gaussian mixture posterior shown in Figure~\ref{fig:2comp_posterior}). We get a mixture of 4's and 9's corresponding to the two posterior peaks. Right panel: samples overplotted with the mask demonstrating that they are compatible with the input data within the errors of the generative model.}
    \label{fig:corr_recon}
\end{figure}  
To demonstrate the methodology, we consider several examples of data corruption on the MNIST dataset \citep{LecunMNIST}. We begin by training a VAE on the uncorrupted training set following a setting similar to \citep{GritsenkoSnoekEtAl19} with both encoder, $f_\psi$, and decoder, $g_\phi$, parameterized as a sequence of 4 fully-connected ResNet blocks, each block containing 2 fully-connected layers with size 512 and LeakyRelu activation. The encoder reduces the dimensionality to 10. We minimize the loss using ADAM~\citep{KingmaB14} with default parameters, a batch size of 1024, and a decreasing learning rate starting at 0.001. 

\setlength{\intextsep}{1 pt}
\begin{wrapfigure}{R}{0.3\linewidth}
\vspace{-3pt}
\begin{center}
    \includegraphics[trim={0.5cm 1cm 0.5cm 1cm},clip,width=0.49
    \linewidth]{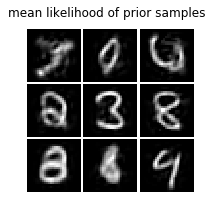}
    \includegraphics[trim={0.5cm 1cm 0.5cm 1cm},clip,width=0.48
    \linewidth]{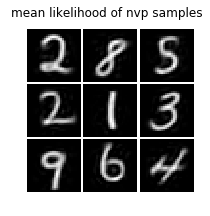}
\end{center}
\vspace*{-2mm}
\caption{Forward modeled samples from the prior before (left) and after (right) augmenting the forward model with a normalizing flow.}
\label{fig:prior_samples}
\end{wrapfigure}
Training a VAE with the ELBO as objective function is not guaranteed to result in an encoded distribution that perfectly matches the prior. This manifests itself in poor sample qualities when forward modeling samples drawn from the prior distribution. A number of approaches have been suggested to improve on this; most of them modify the objective function~\citep{Makhzani2015,Mescheder2017,Tolstikhin2017,Ulyanov2017,Beitler2018}. Here we choose a simple, different approach: to ensure that the prior is well described by a unit variance Gaussian, we augment the forward model by a normalizing flow (specifically, we use a RealNVP~\citep{DinhSB16}). The normalizing flow is trained to map the latent space distribution of the VAE to a standard normal distribution. The full forward model, i.e. the mapping from latent space to data space then simply becomes a successive application of the generator of the normalizing flow followed by the generator of the VAE. The entire procedure is also illustrated in Appendix~\ref{app:fwdmodel}. In Figure~\ref{fig:prior_samples} we show how the samples from the generative model improve after this augmentation. Adding an additional flow model might not be required for all data sets and generative models. If it is required can be judged from the sample quality. In our example, the latent space distribution is already close to a Gaussian. In this case, a RealNVP with eight affine coupling layers, each with hidden layers consisting of [512,512] units, is sufficient to achieve almost perfect samples. 

\setlength{\intextsep}{3 pt}
\begin{wrapfigure}{l}{0.3\textwidth}
    \vspace{-3pt}
    \centering
    \includegraphics[trim={0 0.04cm 0 0.09cm},clip,width=1
    \textwidth]{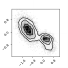}
    \caption{Samples from a Gaussian mixture model fitted to the latent space posterior of the example in the last column of the left panel of Figure~\ref{fig:corr_recon}. We plot a projection of the 10 dimensional latent space to 2 dimensions.}
    \label{fig:2comp_posterior}
\end{wrapfigure}
We then use our reconstruction technique on corrupted images produced from the MNIST test data, shown in the top row of the left panel in Figure~\ref{fig:corr_recon}. In the first we randomly remove 95\% of the pixels~\citep{DalcaEtAl19}, in the second we mask half of the image and add noise with $\sigma_n{=}0.5$ to the other half, in the third we blur the image by convolution with a Gaussian kernel.  In the last example we apply a broad mask, such that different digits (4s and 9s) are compatible with the data.

To reconstruct the images, we perform the posterior analysis outlined in the previous section. We first use optimization to find all local minima of the negative log posterior: we start from different random positions drawn from the prior and use the ADAM optimizer to descend to the local minima. We stop our search once we find that the minimization procedure repeatedly converges to the same minima, typically after \textasciitilde10 minimizations. We plot reconstructions from the lowest minimum in the middle row of the left panel of Figure~\ref{fig:corr_recon} and the true input image for comparison below them. An example of a latent space posterior is shown in Figure~\ref{fig:2comp_posterior}: there are two posterior peaks that dominate the posterior mass, with the first one containing 25\% and the second one containing 75\% of the posterior mass. 

We represent uncertainty quantification in data space by drawing samples from the latent space posterior, and evaluating them in data space using the generator. In the example of the masked 4 there is a range of possible solutions allowed given the broad and bi-modal posterior (Figure~\ref{fig:2comp_posterior}): some of the drawn digits are closer to a 4 while others are closer to a 9. We plot unmasked and masked samples in the two right panels of Figure~\ref{fig:corr_recon} to show that the unmasked data is reproduced within the errors. We provide code to reproduce our experiments in a public github repository \href{https://github.com/bccp/DeepUQ}{\faGithub}.

{\bf Related work:}
\label{sec:related_work}
Deep learning approaches to inverse problems \citep[e.g.][]{DongLHT15, JinMcCann2016, RIMPutzky, DeepImagePrior}, can return high quality point estimates, but usually do not provide uncertainty estimates, which are crucial to many applications (e.g, medical imaging or scientific applications). Data imputation with VAEs was initially suggested by~\cite{RezendeMW14} and further refined by~\cite{MIWAE,Mattei2018}. These methods rely on expensive sampling to marginalize over $\bi{z}$ and evaluate $p(\bi{x}|\bi{y})$. An approach similar to this work was recently developed by ~\cite{WuDomkeEtAl18}, but uses SVI for fitting the posterior model, which can be orders of magnitude slower than the EL$_2$O procedure used in this work. An alternative approach to uncertainty quantification based on learning to sample from the posterior with a W-GAN was proposed by~\cite{2018AdlerOzan}. Their approach requires retraining for different corruption types, and should be affected by the well known limitations of GANs (training instability and mode collapse). 

\section{Conclusions}
Uncertainty quantification of complex and strongly corrupted data can be challenging, and Bayesian posterior analysis 
is an approach that can quantify complex posterior distributions such as multimodal solutions. It 
requires a reliable prior distribution in latent space, which can be 
obtained using generative models such as a VAE. Posterior analysis with
MCMC is often too slow and can be trapped in a single mode. We propose a fast alternative based on the L$_2$ f-divergence that gives high quality posteriors in latent space and in data space.

\newpage
\bibliography{cosmo,neurips,vae_sample_improvements}  
\bibliographystyle{icml2019}

\appendix

\section{Details on the Posterior Analysis}
\label{app:el2o}
\subsection{The EL$_2$O procedure}
EL$_2$O~\citep{SeljakYu19} is an efficient and numerically stable procedure for fitting a parameterized distribution $q_{\theta}(\bi{z})$ to another distribution $p(\bi{z})$, when we can evaluate $p(\bi{z})$ for a given $\bi{z}$, but do not have access to the analytical form of the distribution. In its most general form EL$_2$O determines the parameters, $\theta$, of $q_\theta$ by minimizing the following objective function,
\begin{equation}
    \label{eq:genEL2O}
    \EO= \underset{\theta}{\mathrm{arg min}}\, \mathbb{E}_{\tilde{p}} \Big \{ N^{-1} \sum_{n=0}^{n_\mathrm{max}} \sum_{i_1,...,i_n} \alpha_n \left[ \nabla^n_{\bi{z}} \ln q_\theta(\bi{z})-\nabla^n_{\bi{z}} \ln p(\bi{z}) \right]^2 \Big \},
\end{equation}
where the expectation value is taken by averaging over samples from an arbitrary distribution $\tilde{p}$ (ideally $\tilde{p}$ is chosen to be close to the real distribution $p(\bi{z})$), $N$ is the number of terms included, and the factor $\alpha_n$ can be used to give different weight to different derivative orders. The sum over indices $i_1,..i_n$ should be symmetrized to avoid double counting.

Setting $n_{\max}=2$, Eq.~\ref{eq:genEL2O} becomes
\begin{align}
    \nonumber
    \EO= N_M^{-1}\, \underset{\theta}{\mathrm{arg min}}\,  \mathbb{E}_{\tilde{p}} \Big \{ & \sum_{i,j<i}^{M} \left[ \nabla_{z_i} \nabla_{z_j}\ln q_\theta(\bi{z})- \nabla_{z_i} \nabla_{z_j} \ln p(\bi{z}) \right]^2 \\
    \nonumber
    & + \sum_{i=1}^{M} \left[ \nabla_{z_i} \ln q_\theta(\bi{z})- \nabla_{z_i} \ln p(\bi{z}) \right]^2\\
    \label{eq:EL2O}
    & +  \left[ \ln q_\theta(\bi{z})- \ln p(\bi{z}) - \ln \bar{p}\right]^2 \Big \},
\end{align}
with $N_M=M(M+ 3)/2 + 1$. Note that we have added a normalization, $\ln \bar{p}$, in the last term to account for the fact that $p(\bi{z})$ might not be properly normalized. The last term is only required if the correct normalization for $q_\theta$ has to be determined, e.g. if $q_\theta$ is a mixture of Gaussians and their relative weights have to be found.

Specializing to the Gaussian case, $q_\theta(\bi z)=\mathcal{N}\left(\bi{z};\bi{\mu},\bi{\Sigma}\right)$, we have $\theta=(\bi{\mu}, \bi{\Sigma})$. For $N_k$ random sampling points, $\bi{z_k}$, the $\EO$ is minimized by setting
\begin{equation}
    \bi \Sigma^{-1}\approx -N_k^{-1} \sum_{k=1}^{N_k} \nabla_{\bi z} \nabla_{\bi z} \ln p(\bi z_k),
\end{equation}
and
\begin{equation}
    \bi \mu \approx  N_k^{-1} \sum_{k=1}^{N_k}\left[\bi \Sigma \nabla_{\bi z} \ln p(\bi z_k)+ \bi{z}_k \right].
\end{equation}
This results in a first estimate of $q_\theta$, which can be used to draw new samples, $\bi{z}_k$, for which the fitting procedure can be repeated, etc.  $\EO$ and variational inference by minimization of the KL-divergence, $\mathrm{KL}(q|p)$, converge to the same result in the high sample limit, if $\tilde{p}=q$.  However, $\EO$ has the advantage of lower sampling noise: the presence of $\bi{z}_k$ in the last equation guarantees that there is no sampling noise. This term is missing in stochastic KL minimization \citep{L2Convergence}. 

We can further accelerate the convergence by choosing the initial sampling points wisely, e.g. for a unimodal distribution, one can first find the maximum, $\bi{\tilde{z}}$, of $p(\bi{z})$ through gradient descent. If one chooses $\bi{z}_k=\bi{\tilde{z}}$, $\EO$ simplifies to the well known Laplace approximation.

Since we expect the posterior of corrupted data to be generally multimodal, we use $\EO$ to fit a Gaussian mixture model (GMM), $q_\theta=\sum_i w_i q_{i \theta}=\sum_i w_i \mathcal{N} \left(\bi{z};\bi{\Sigma}_i,\bi{\mu}_i\right)$. Assuming that the distribution $p(\bi{z})$ has well separated maxima, $\tilde{\bi{z}}_i$, finding the parameters of the GMM proceeds similarly as for the single Gaussian case. In fact, we can simply fit Gaussians as outlined above around the maxima. In addition, we need to determine their relative weights, $w_i$. The additional constraint comes from the last term in Eq.~\ref{eq:EL2O} (which was not required to find the parameters of the single Gaussian). We see that $\EO$ gets minimized if we set
\begin{equation}
    \ln(w_i)= -\ln q_i(\tilde{\bi{z}}) + \ln p(\tilde{\bi{z}}) + \ln \bar{p},
    \label{eq:weights}
\end{equation}
and require $\sum_i w_i=1$.

\subsection{Posterior Analysis in Practice}
\label{app:post_analysis}
\begin{wrapfigure}{R}{0.5\textwidth}
    \vspace{-3pt}
    \centering
    \includegraphics[trim={0 0.04cm 0 0.09cm},clip,width=1
    \textwidth]{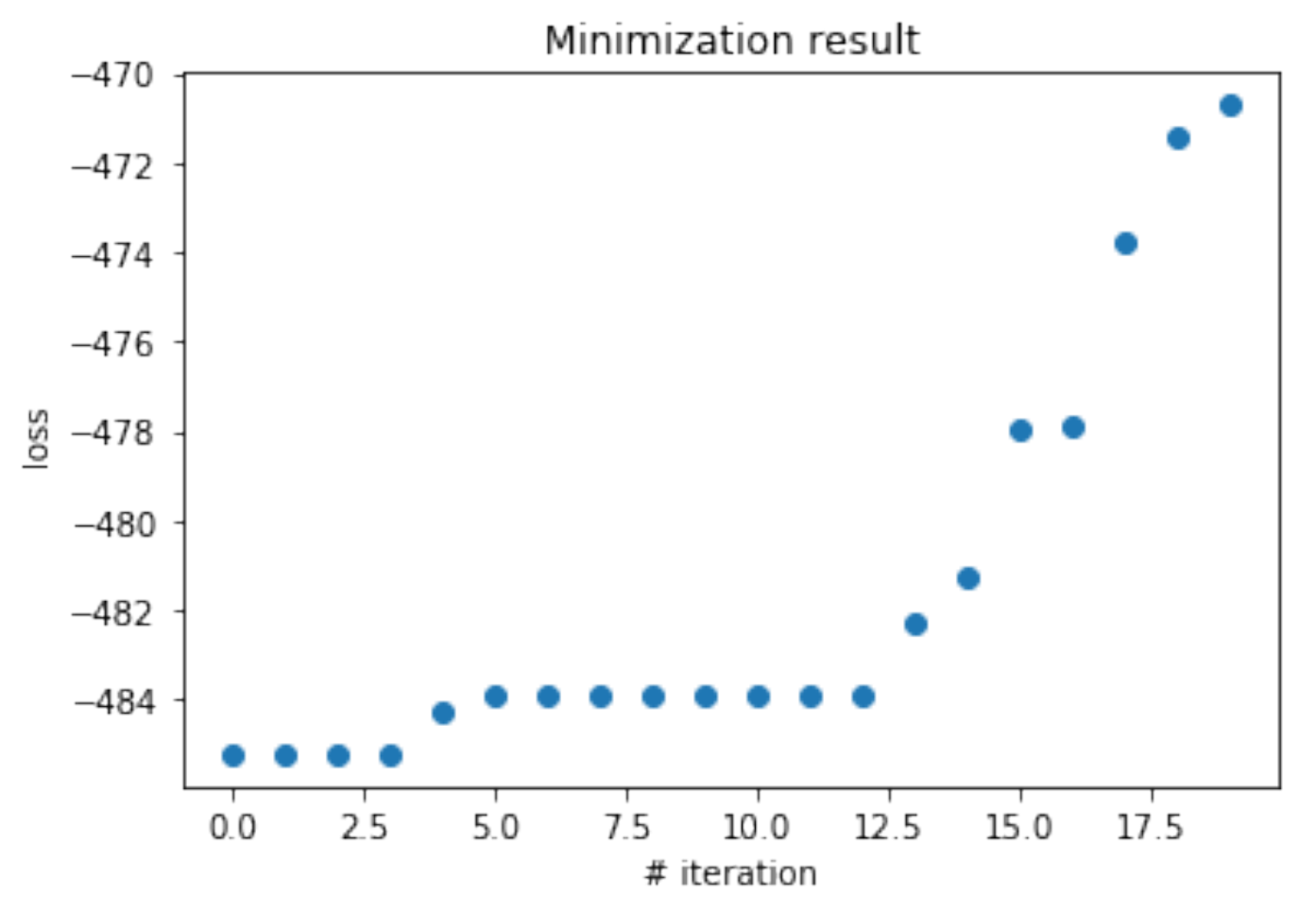}
    \caption{Outcomes (negative log posterior) of 20 minimizations starting from 20 different, random starting points, ordered by their negative log posterior value. The two dominating minima (c.p. Fig.~\ref{fig:2comp_posterior}) are found reliably. The other minima are found to contribute negligibly to the total posterior mass (The $\EO$ prcedure assigns them a very small weight, c.p. Eq.~\ref{eq:weights}).}
    \label{fig:minimization_results}
    \vspace*{-2mm}
\end{wrapfigure}

We use the fitting procedure outlined in the previous section for the examples shown in the main text. First, we determine all the relevant minima by running several gradient descents with decreasing learning rate from random starting points that we draw from the prior distribution. Each minimization only takes of the order of 10 seconds and we find that we rarely need to start from more than 10 different starting points to find all the relevant minima. Once the gradient descent has converged, we check the Hessian, $-\nabla_{\bi z} \nabla_{\bi z} \ln p(\bi z)$, at the endpoint for positive definiteness. This step ensures that we do not mistake saddle points as minima. We have found that the minimization is less likely to get stuck at points other than actual minima if we apply an annealing scheme in which we initially downweight the likelihood compared to prior. An example of the outcomes of twenty minmizations from different starting points for the example of the masked 4 (last column in Figure~\ref{fig:corr_recon}), which has the most complex posterior of all our examples, is shown in  Figure~\ref{fig:minimization_results}.

 We can clearly see that the minimization reliably finds one of the two deepest minima. Note that following $\EO$ the relative importance of minima (their relative weights) does depend their posterior mass, which is determined by both their depth and width. A deep, but very local minimum might have less posterior mass than a shallower, wide minimum. Our minimization procedure is unlikely to miss wide minima and in all examples, we find that the forward modeled lowest minimum is always very close to the input, suggesting that it is the global minimum.

After minimization and discarding all points with non-positive definite Hessian, we identify separate minima, $\tilde{\bi z}_i$, by comparing their distance in latent space $|\tilde{z}_{i_n}{-}\tilde{z}_{j_n}|$ (per latent space dimension $n$) to their width (the width is estimated from the variance, $\sigma_{i_n}^2{=} \nabla_{z_n} \nabla_{z_n} \ln p(\tilde{\bi z}_i)$, at the respective minima). Having identified all separate minima, we fit Gaussian around them and determine their relative weights with Eq.~\ref{eq:weights}.

\subsection{ EL$_2$O versus stochastic variational inference (SVI)}
\label{app:elosvi}
\begin{figure}
    \vspace{-3pt}
    \centering
    \includegraphics[width=0.15\textwidth]{truth_masknoise05.pdf}
    \includegraphics[width=0.15\textwidth]{input_data_masknoise05.pdf}
    \includegraphics[width=0.15\textwidth]{lowest_minimum_masknoise05.pdf}
    \includegraphics[width=0.15\textwidth]{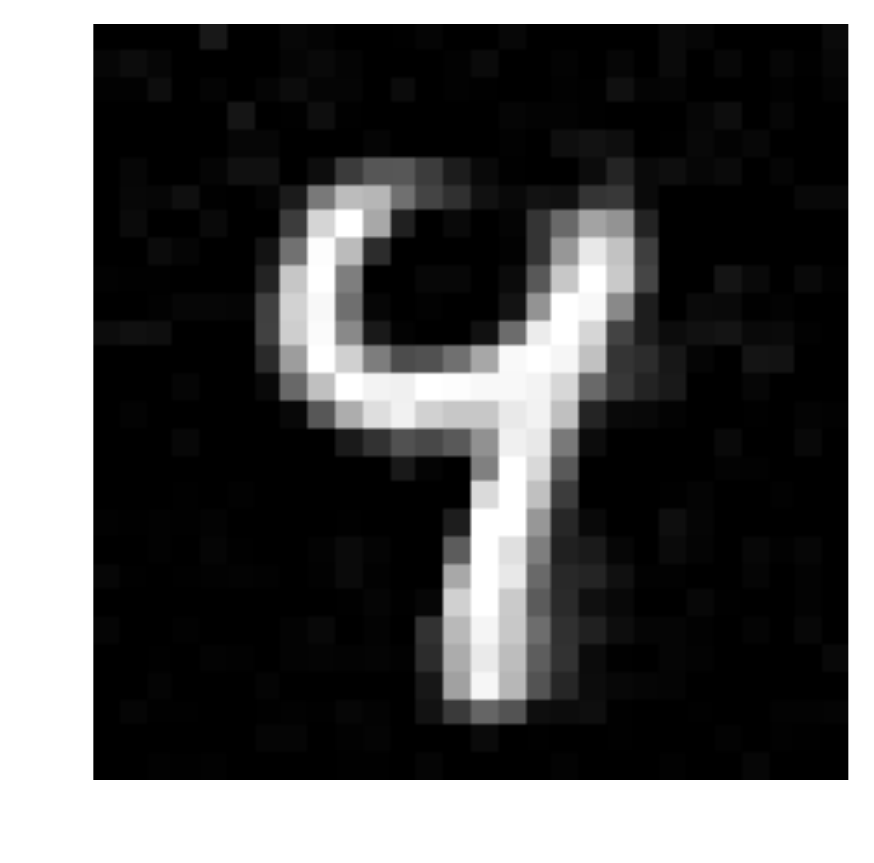}
    \includegraphics[width=0.15\textwidth]{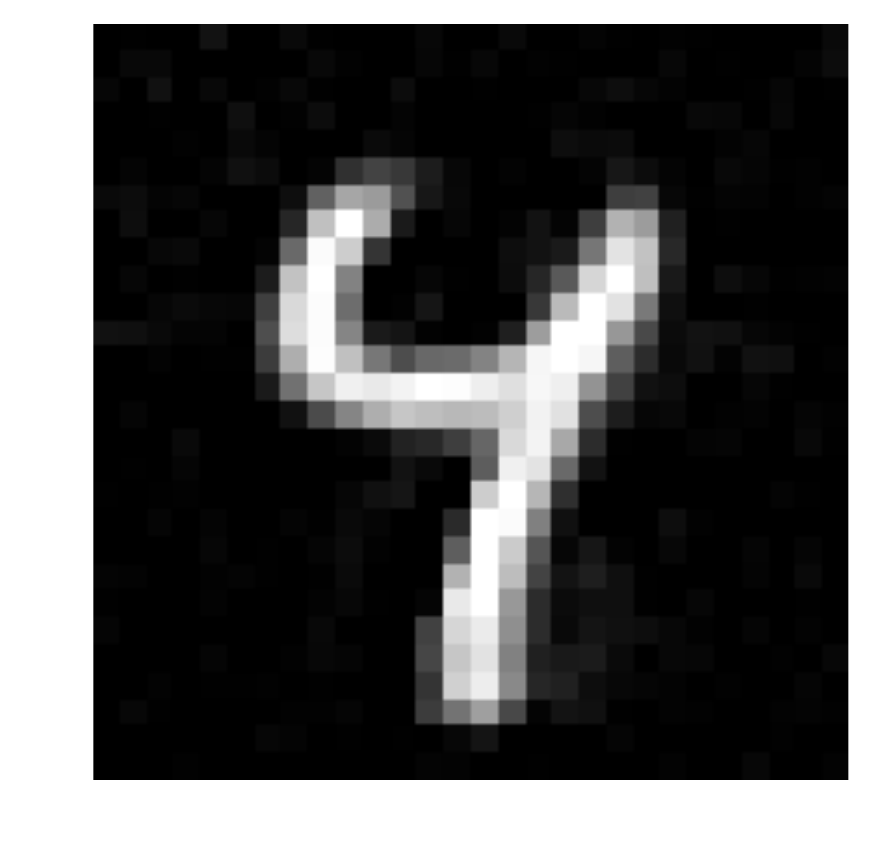}
    \caption{Comparison of different posterior fitting procedures on the example of the noisy, half masked four. In the two images on the left, we show the true underlying image and the input data, which has been corrupted by adding white noise and a mask covering half of the image. In the right three images we show the maxima of the fitted Gaussians forward modeled to data space, starting with the $\EO$ estimate (MAP solution), followed by the mean field VI estimate and the full rank VI estimate on the right. The VI estimates are slightly different to the $\EO$ estimate, reflecting the fact that their maxima do not coincide with the global maximum (c.p. Fig.~\ref{fig:probed_posterior}).}
    \label{fig:noise_only}
    \vspace*{-2mm}
\end{figure}
\begin{figure}[h!]
    \includegraphics[width=0.99\textwidth]{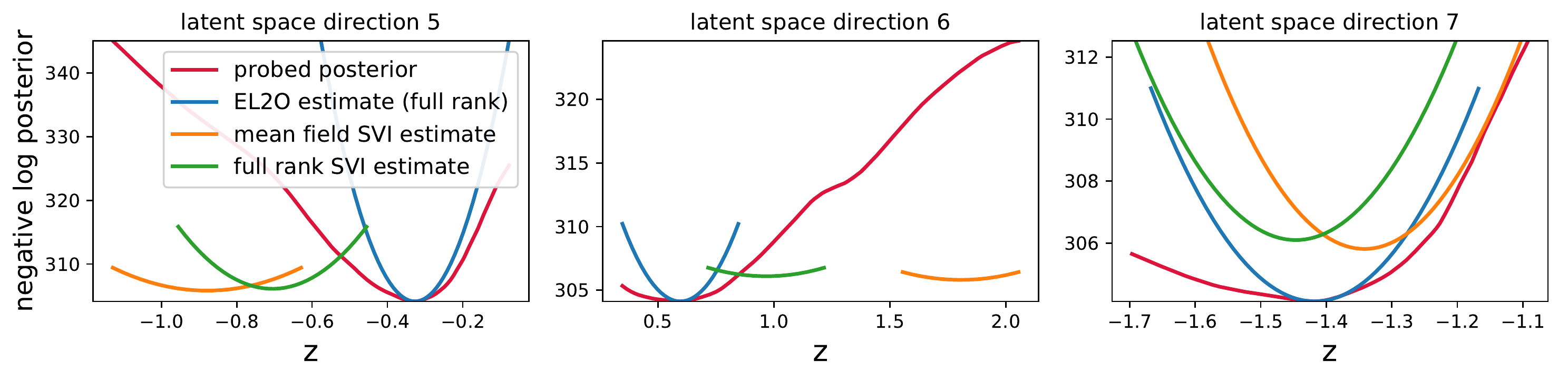}
    \caption{A comparison of posterior fits obtained with different fitting procedures to the true posterior. We show 3 randomly chosen latent space directions of the 10 dimensional latent space and plot the negative log posterior. We find that minimizing the ELBO requires more fine tuning than minimizing the $\EO$ objective, while not leading to fitted posterior modes that coincide with the true global mode.}
    \label{fig:probed_posterior}
\end{figure}

We compare different posterior fitting procedures for the second example in Figure~\ref{fig:corr_recon}, for which we found the posterior to be to good approximation unimodal. In this case, $\EO$ reduces to finding the minimum and fitting a full rank Gaussian by means of the Laplace approximation. For comparison, we also fit Gaussians with diagonal (mean field approximation) and full rank covariance to the posterior by stochastically minimizing the evidence lower bound (ELBO)~\citep{KucukelbirTRGB17}. For $\EO$ we run 10 minimization starting from random starting points, each taking about 10 seconds. Out of these ten minimizations, 4 converge to the deepest minimum, and we find that the contribution of the other local minima to the posterior is suppressed. We compute the curvature at the global minimum by automated differentiation and use its inverse as an estimate of the covariance of our Gaussian fit. For the stochastic VI estimate we minimize the ELBO using the ADAM optimizer with decreasing learning rate. We draw 512 samples from the approximate posterior to estimate the gradient in each step (we see a clear trend of improved results with increased number of samples). We find that we need of the order of 8000 steps for the ELBO to converge, taking between 130 (mean field) and 170 (full rank) seconds, making it in both cases slower than running 10 minimizations for the $\EO$ estimate. Despite using the Cholesky decomposition and ensuring that the diagonal is positive with a softplus transform, we find that we have to add a small constant offset to the diagonal of the estimated covariance to make the minimization of the ELBO numerically stable.
We plot the outcomes of these fits in Figure~\ref{fig:probed_posterior}. The red line shows the true negative log posterior probed along one specific latent space dimension (keeping the other dimensions fixed at the position of the global minimum), the other lines show the respective estimates from the fitted posteriors, $q_\theta(\bi{z})$, restricted to the same dimension. We use the value of the true negative log posterior evaluated at the mean of the fitted posterior as anchor point for the fits (which is why the minima of the SVI estimates do not need to lie on the red line). The minima of the negative log posteriors fitted with the ELBO objective do not coincide with the true mininum. The reconstructions from these points have less similarity with the truth than the MAP estimate (Fig.~\ref{fig:noise_only}), the fact that they still look similar to the MAP solution is probably owed to the relative simplicity of the posterior shape, but could become more problematic for other posterior shapes.

\section{A detailed description of the forward model}
\label{app:fwdmodel}
\begin{figure}[h!]
    \includegraphics[width=0.5\textwidth]{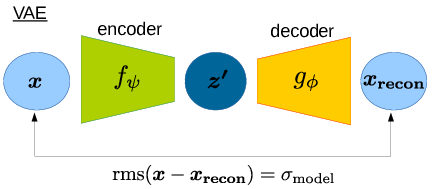}
    \caption{As a first step we train a vanilla VAE with Gaussian likelihood on uncorrupted training data.}
    \label{fig:networks1}
\end{figure}

\begin{figure}[h!]
    \includegraphics[width=0.5\textwidth]{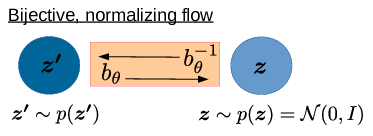}
    \caption{In a second step a bijective normalizing flow is fitted to map the distribution of the encoded training data to a normal distribution.}
    \label{fig:networks2}
\end{figure}

\begin{figure}[h!]
    \includegraphics[width=0.9\textwidth]{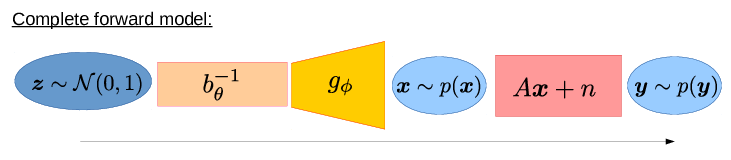}
    \caption{The full forward model consists out of the normalizing flow, the decoder of the VAE, the corruption operator and measurement noise.}
    \label{fig:fwdmodel}
\end{figure}

As described in the main text, we start by training the encoder and decoder of a VAE under the evidence lower bound (Figure~\ref{fig:networks1}), assuming a Gaussian likelihood. The modeling error of this VAE can be estimated from the root mean squared (rms) difference between the input and the reconstruction or by allowing the variance of the likelihood to be a trainable variable and using its value at the end of training as an estimate.
To ensure that the latent space distribution is well described by the prior, we add an additional component to the forward model: We first encode the data with the VAE encoder. We then fit a bijective normalizing flow, $b_\theta$, to map this distribution to a normal distribution (Figure~\ref{fig:networks2}). 
After having trained both the VAE and the normalizing flow, we build our complete forward model ( Figure~\ref{fig:fwdmodel}): The hidden variable $\bi{z}$ is passed through the bijective flow and the decoder of the VAE to produce a realization of the uncorrupted data $\bi{x}$. This data is then corrupted by the operator $\bi A$ and a realization of the measurement noise is added. This produces a realization of noisy corrupted data. 
\end{document}